\documentclass[11pt]{article}

\usepackage{acl}
\usepackage{times}
\usepackage{latexsym}
\usepackage[T1]{fontenc}
\usepackage[utf8]{inputenc}
\usepackage{microtype}
\usepackage{inconsolata}

\usepackage{amsmath}
\usepackage{amssymb}
\usepackage{pifont}
\usepackage{multirow}
\usepackage{multicol}
\usepackage{booktabs}
\usepackage{graphicx}
\usepackage{threeparttable}
\usepackage{enumitem}

\usepackage{todonotes}

\begin{document}
\title{Less Is More? Selective Visual Attention to High-Importance Regions for Multimodal Radiology Summarization}

\author{
\textbf{Mst. Fahmida Sultana Naznin\textsuperscript{1}},
 \textbf{Adnan Ibney Faruq\textsuperscript{1}},
 \textbf{Mushfiqur Rahman\textsuperscript{1}},
 \\
 \textbf{Niloy Kumar Mondal}\textsuperscript{1},
 \textbf{Md. Mehedi Hasan Shawon\textsuperscript{2}},
 \textbf{Md Rakibul Hasan\textsuperscript{3}}
\\
\\
 \textsuperscript{1}Bangladesh University of Engineering and Technology, Dhaka 1000, Bangladesh
\\
 \textsuperscript{2}BRAC University, Dhaka 1212, Bangladesh
\\
 \textsuperscript{3}Curtin University, Bentley, WA 6102, Australia
\\
 \small{
   \textbf{Correspondence:} \href{mailto:rakibul.hasan@curtin.edu.au}{rakibul.hasan@curtin.edu.au}
 }
}

\maketitle

\begin{abstract}
Automated radiology report summarization aims to distill verbose findings into concise clinical impressions, but existing multimodal models often struggle with visual noise and fail to meaningfully improve over strong text-only baselines in the FINDINGS $\to$ IMPRESSION transformation. We challenge two prevailing assumptions: (1) that more visual input is always better, and (2) that multimodal models add limited value when findings already contain rich image-derived detail. Through controlled ablations on MIMIC-CXR benchmark, we show that selectively focusing on pathology-relevant visual patches rather than full images yields substantially better performance. We introduce \textbf{ViTAS}, \textbf{V}isual-\textbf{T}ext \textbf{A}ttention \textbf{S}ummarizer, a multi-stage pipeline that combines ensemble-guided MedSAM2 lung segmentation, bidirectional cross-attention for multi-view fusion, Shapley-guided adaptive patch clustering, and hierarchical visual tokenization feeding a ViT. ViTAS achieves SOTA results with 29.25\% BLEU-4 and 69.83\% ROUGE-L, improved factual alignment in qualitative analysis, and the highest expert-rated human evaluation scores. Our findings demonstrate that \emph{less but more relevant} visual input is not only sufficient but superior for multimodal radiology summarization. 
% Code is publicly available at TBA.
\end{abstract}
% superior CheXbert and RadGraph

\section{Introduction}

Radiology report summarization is essential for distilling verbose diagnostic findings into concise, actionable impressions, thereby supporting clinical decision-making and reducing radiologist workload~\cite{miura-etal-2021-improving}. With the rapid growth of medical imaging volumes, automated multimodal summarization-integrating both textual findings and radiological images-has emerged as a promising direction to produce more clinically faithful and contextually grounded impressions~\cite{kim-etal-2023-ku, wang-etal-2023-utsa, nicolson-etal-2023-e, delbrouck-etal-2021-qiai, zhang2024leveraging}.

Two influential assumptions have shaped recent research in this area. First, many works implicitly follow a “the more, the better” heuristic: the more visual input provided to the model (full images, multiple views, higher resolution), the better the summarization performance~\cite{wang-etal-2023-utsa}. This assumption has rarely been systematically challenged in the radiology domain, even though irrelevant regions frequently introduce visual noise that can dilute cross-modal alignment and degrade factual accuracy~\cite{degrave2021ai, rajaraman2024noise, niehoff2023evaluation}.
Second, several studies argue that multimodal models add limited value when transforming richly detailed FINDINGS sections (already written by radiologists with comprehensive image interpretation) into concise IMPRESSION sections. According to this view, if the FINDINGS section contains sufficient detail, it should already encode all key radiographic information, and the primary limitation of multimodal models lies in their failure to learn meaningful cross-modal interactions between text and visual representations~\cite{sim-etal-2025-more}.

We present the first study on MIMIC-CXR that directly tests \emph{both} assumptions in the same controlled setting. Our results show that: contrary to the first assumption, feeding full uncurated images (even multi-view) is \emph{not} optimal-selective incorporation of only the most pathology-relevant visual patches yields substantially larger gains than simply increasing the amount of visual input; contrary to the second assumption, properly integrated multi-view visual context \emph{does} meaningfully improve FINDINGS $\to$ IMPRESSION summarization quality across all major automatic and clinical metrics. To achieve this, we propose ViTAS, a multi-stage framework that systematically removes noise and focuses the model on clinically salient information. The pipeline begins with robust ensemble-guided lung segmentation using MedSAM2~\cite{ma2025medsam2}, followed by dual-backbone Swin Transformer V2 processing with bidirectional spatial cross-attention for multi-view fusion. Post-hoc gradient-weighted attention maps combined with Shapley value-based view contribution quantification enable dynamic, pathology-adaptive selection of high-importance patch clusters. These selected clusters are then hierarchically tokenized (global + cluster + dynamic extra patches) and fused with textual findings in a ViT + T5 multimodal transformer to generate structured impressions. Our main contributions are:
\begin{itemize}
    \item We introduce an end-to-end pipeline combining ensemble MedSAM2 lung segmentation, bidirectional mid-level cross-attention, view-adaptive patch clustering, and hierarchical visual tokenization. The approach achieves SOTA automatic metrics, substantially improves factual alignment and reduces hallucination risk by qualitative analysis, and obtains the highest expert-rated scores in human evaluation across readability, factual correctness, informativeness, and completeness.
     \item We show that selective, attention- and Shapley-guided patch extraction outperforms both full-image and naive ROI-based multimodal baselines, providing the first clear evidence in radiology summarization that \emph{less (but more relevant) visual input is better}.
    \item We demonstrate, through controlled ablations on MIMIC-CXR, that multimodal models \emph{can} significantly outperform strong text-only baselines in the FINDINGS $\to$ IMPRESSION task when visual noise is effectively removed-challenging the view that multimodal gains are inherently limited in previous setting.

\end{itemize}

\section{Related Work}
With the rapid advancement of multimedia content, multimodal summarization has gained increasing attention to generate concise and informative summaries~\cite{1li-etal-2018-multi, 2jangra-etal-2021-multi, 3overbay-etal-2023-mredditsum, 4liang-etal-2023-summary, 5sanabria-etal-2018-how2, 6mahasseni-etal-2017-unsupervised, 7liu-etal-2024-multimodal, 8mukherjee-etal-2022-topic, 9yu-etal-2021-vision}. Recent studies have begun to explore incorporating radiology images as inputs to multimodal summarization models, under the assumption that their rich visual information can enhance the quality of generated summaries~\cite{ghosh-etal-2024-sights, thawakar-etal-2024-xraygpt, hu-etal-2023-improving, van-veen-etal-2023-radadapt, chen-etal-2023-toward, delbrouck-etal-2023-overview}. Some works focus on pretraining vision-language models such as CLIP~\cite{radford2021learning} and ViT~\cite{dosovitskiy2020image}, while others address radiology report summarization from findings sections~\cite{zhang-etal-2018-learning-summarize} or explore multimodal summarization techniques including guided attention across modalities~\cite{atri2021see}, self-supervised opinion summarization with images~\cite{im-etal-2021-self}, and multimodal approaches specifically for radiology reports~\cite{delbrouck-etal-2021-qiai}.

Several studies have explored modality contributions in multimodal models~\cite{goyal-etal-2017-making}. \citeauthor{parcalabescu-frank-2023-mm}(\citeyear{parcalabescu-frank-2023-mm}) employ Shapley values by randomly masking image patches and text tokens to quantify changes in output. \citeauthor{liang-etal-2023-quantifying}(\citeyear{liang-etal-2023-quantifying}) introduced an information-theoretic framework to measure multimodal interactions in datasets and performed human annotation studies on general-domain data~\cite{liang-etal-2023-multimodal}.

CSTRL~\cite{naznin-etal-2025-cstrl} shows that learning from salient textual statements improves summarization, highlighting that not all contextual information is equally informative. While textual saliency has been explored, selective visual importance in multimodal radiology summarization remains under-investigated. Although NEURAL~\cite{joshi-etal-2025-neural} and MSCL~\cite{zhao-etal-2023-mscl} demonstrate the effectiveness of salient visual regions for report generation, most multimodal summarization approaches still assume a “the more, the better” paradigm or suggest that textual information alone can often be sufficient.

\begin{figure*}[!ht]
\centering
\includegraphics[width=\textwidth]{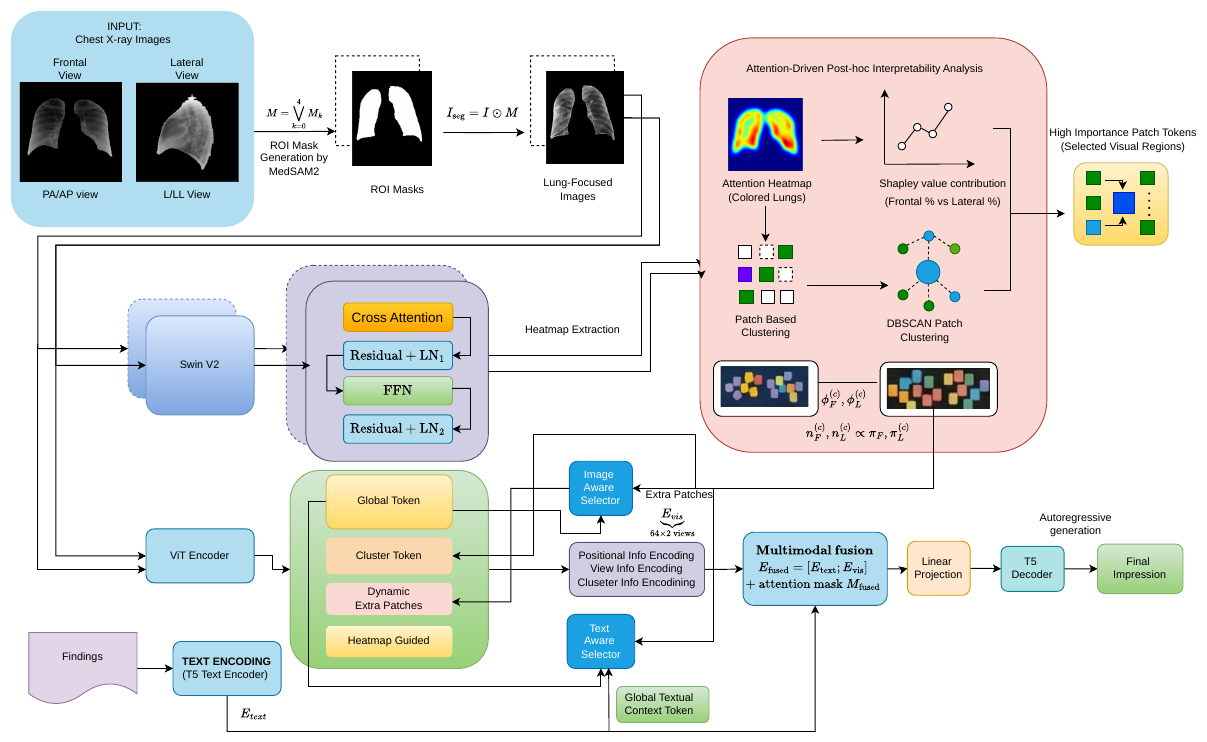}
\caption{Overview of the \textbf{ViTAS} pipeline: Chest X-rays (frontal and lateral) are lung-segmented using MedSAM2, then fused via a dual SwinV2 with cross-attention. Interpretability modules identify important regions, which are selectively tokenized and combined with text embeddings in a T5 decoder to generate the final clinical impression.
}
\label{fig:vit as_overview}
\end{figure*}

\section{Method}
We propose a multi-stage framework for radiology report summarization that prioritizes clinically relevant information while discarding irrelevant regions. First, lung regions are accurately segmented using an ensemble-guided MedSAM2 approach to focus the model on pulmonary anatomy. Frontal and lateral views are then processed through a dual-backbone Swin Transformer V2 architecture with bidirectional spatial cross-attention, enabling complementary information exchange while preserving view-specific features. Attention-driven post-hoc analysis identifies high-importance patches and quantifies each view’s contribution using Shapley values, allowing dynamic selection of the most informative regions. Finally, the selected patch clusters are fused with textual prompts in a multimodal transformer framework (ViT + T5) to generate structured clinical impressions. An overview is provided in Figure~\ref{fig:vit as_overview}.

\subsection{ROI-Guided Visual Context Extraction}

\begin{figure}[!t]
\centering
\includegraphics[width=0.5\textwidth]{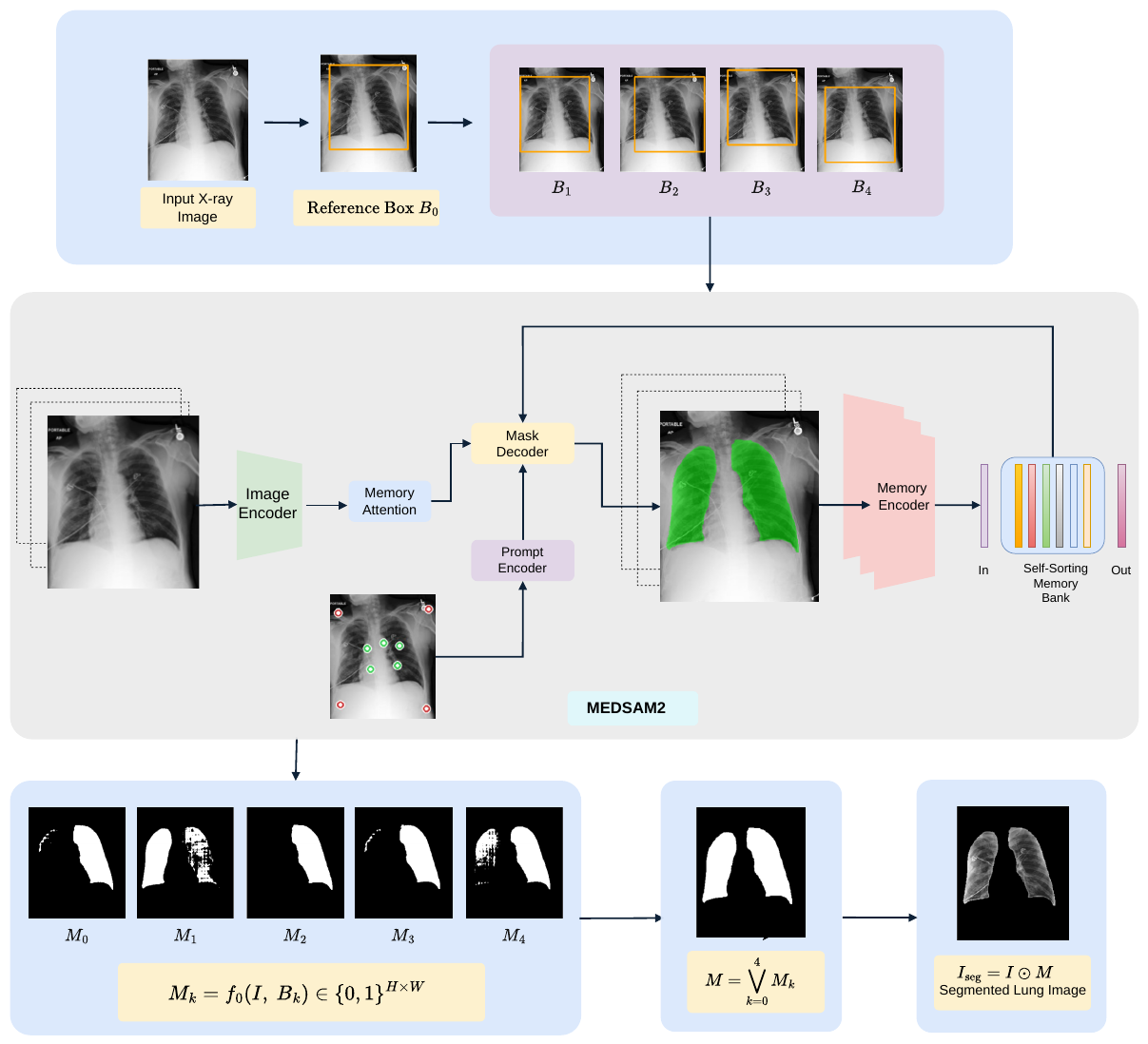}
\caption{Ensemble-guided MedSAM2 lung segmentation pipeline. A reference box generates five shifted bounding boxes. MedSAM2 produces binary masks which are unioned into the final lung ROI mask. The segmented image retains only pulmonary fields.}
\label{fig:medsam2_ensemble}
\end{figure}

Let $I \in \mathbb{R}^{H \times W \times 3}$ denote a chest X-ray image with height $H$ and width $W$ and $\mathcal{V} \in \{\text{AP}, \text{PA}, \text{L}, \text{LL}\}$ indicate its projection view (Anteroposterior, Posteroanterior, Lateral, Left Lateral, respectively). We process each view independently to avoid view-specific confounding effects as shown in Figure~\ref{fig:medsam2_ensemble}.

\paragraph{Ensemble Bounding-Box Prompting.}

We utilize MedSAM2 model~\cite{ma2025medsam2} with predictor $f_{\theta}$ to obtain binary lung segmentation masks. A reference bounding box $B_0 = [x_1, y_1, x_2, y_2]$ is defined in pixel coordinates using empirical statistics computed from a small set of manually annotated images. Its corners are expressed in normalized image coordinates as:
\(\left( \frac{x_1}{W},\ \frac{y_1}{H},\ \frac{x_2}{W},\ \frac{y_2}{H} \right)\). Here $(x_1, y_1)$ and $(x_2, y_2)$ represent the top-left and bottom-right corners, respectively.

To increase robustness against anatomical variability, patient positioning differences, and model sensitivity to exact box placement, we create an ensemble $\mathcal{B} = \{B_k\}_{k=0}^{4}$ of $K=5$ bounding boxes by applying small directional shifts of magnitude $s$ pixels:
\(
\delta_k \in \{[0,0,0,0],\allowbreak
[-s,0,-s,0],\allowbreak
[+s,0,+s,0],\allowbreak
[0,-s,0,-s],\allowbreak
[0,+s+\Delta,+s+\Delta,0]\},
\)
where $\Delta$ is a small additional downward bias sometimes applied to better cover the lower lung fields (cardiophrenic angles). Each $\delta_k = [\Delta x_1, \Delta y_1, \Delta x_2, \Delta y_2]$ is added element-wise to $B_0$, after which the resulting box coordinates are clipped to lie within $[0, W] \times [0, H]$. For each box $B_k \in \mathcal{B}$, MedSAM2 produces a binary mask:
\begin{equation}
M_k = f_{\theta}(I,\ B_k) \;\in\; \{0,1\}^{H \times W},
\label{eq:single_mask}
\end{equation}
where value 1 indicates a pixel predicted to belong to lung tissue. The final lung region-of-interest (ROI) mask is obtained by taking the pixel-wise logical union (inclusive OR) over the ensemble:
\begin{equation}
M = \bigvee_{k=0}^{4} M_k.
\label{eq:union_mask}
\end{equation}
The union improves robustness to box misalignment and under-segmentation, yielding a more inclusive yet well-localized lung mask.

The segmented lung image is then computed via element-wise multiplication (Hadamard product):
\begin{equation}
I_{\text{seg}} = I \odot M,
\label{eq:segmented_image}
\end{equation}
which suppresses all non-lung regions (setting them to zero) and retains high-fidelity visual context from the pulmonary fields only.

\begin{figure*}[htbp]
\centering
\includegraphics[width=0.8\textwidth]{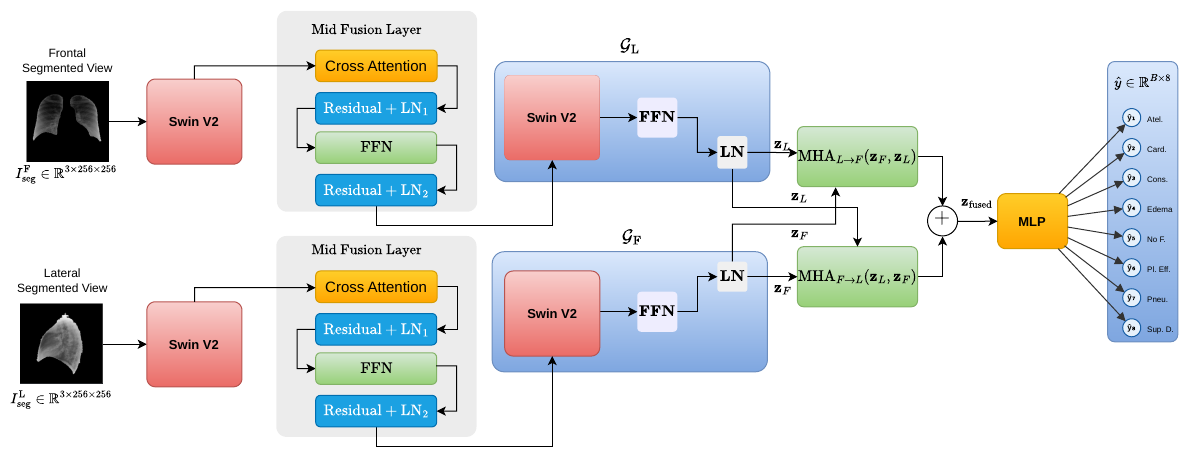}
\caption{Dual Swin Transformer V2 with bidirectional mid-fusion cross-attention. Frontal and lateral views are processed by separate backbones, exchange information via cross-attention, and their global features are fused before classification.}
\label{fig:mid_fusion_architecture}
\end{figure*}
\subsection{Multi-View Feature Fusion with Bidirectional Spatial Cross-Attention}

As shown in Figure~\ref{fig:mid_fusion_architecture}, segmented chest X-ray views are grouped into two complementary categories: frontal (AP, PA) and lateral (L, LL). Lung-focused segmented images
\(
I_{\text{seg}}^{\text{F}}, I_{\text{seg}}^{\text{L}} \in \mathbb{R}^{H \times W \times 3}
\)
are processed by a dual-backbone architecture based on Swin Transformer V2~\cite{Liu2021SwinTV}. Early-stage encoders
\(
\mathcal{E}_{\text{F}}(\cdot),\mathcal{E}_{\text{L}}(\cdot):\mathbb{R}^{3\times256\times256}\rightarrow\mathbb{R}^{N\times D}
\)
extract spatial token sequences from frontal and lateral views, where \(N=(256/p)^2\) is the number of patches and \(D\) is the token dimension. To enable interaction between the two views while preserving view-specific representations, we perform mid-level fusion using bidirectional spatial cross-attention~\cite{zhou2023transformer}. For fusion layer \(\ell\), let
\(
\mathbf{F}^{(\ell-1)},\mathbf{L}^{(\ell-1)}\in\mathbb{R}^{B\times N\times D}
\)
denote frontal and lateral tokens, where \(B\) is the batch size.
Frontal-to-lateral attention is computed as
\begin{equation}
\mathbf{A}_{\text{F}\to\text{L}}=
\text{softmax}\!\Big(\frac{\mathbf{Q}_{\text{F}}\mathbf{K}_{\text{L}}^\top}{\sqrt{d_h}}\odot\mathbf{M}_{\text{L}}\Big),
\end{equation}
where \(h\) is the number of attention heads, \(d_h=D/h\) is the head dimension, and \(\mathbf{M}_{\text{L}}\in\{0,1\}^{B\times1\times1\times N}\) indicates lateral-view availability. The frontal representation is updated as
\begin{equation}
\mathbf{\hat F}^{(\ell)}=\mathbf F^{(\ell-1)}+
\text{Proj}_{\text F}\!\left(\mathbf A_{\text F\to\text L}\mathbf V_{\text L}\right)\odot\mathbf M_{\text L}
\end{equation}
and the lateral update is computed symmetrically. Each branch is refined using residual connections, layer normalization (LN), and a feed-forward network (FFN). To handle missing views, learnable missing tokens
\(
\mathbf m_{\text F},\mathbf m_{\text L}\in\mathbb R^{1\times1\times D}
\)
replace unavailable inputs using view masks.

After \(L\) fusion layers, the token sequences are passed through the remaining Swin stages to produce global descriptors
\(
\mathbf z_{\text F},\mathbf z_{\text L}\in\mathbb R^{B\times D_f},
\)
where \(D_f\) is the final feature dimension.

A lightweight bidirectional cross-attention module produces the fused representation
\begin{equation}
\mathbf z_{\text fused}=
\text{Concat}\Big(
\text{MHA}_{\text L\to\text F}(\mathbf z_{\text F},\mathbf z_{\text L}),
\text{MHA}_{\text F\to\text L}(\mathbf z_{\text L},\mathbf z_{\text F})
\Big).
\end{equation}

A small MLP classifier outputs pathology probabilities
\(
\hat{\mathbf y}\in\mathbb R^{B\times8}
\)
for Atelectasis, Cardiomegaly, Consolidation, Edema, No Finding, Pleural Effusion, Pneumothorax, and Support Devices. The model is trained using binary cross-entropy loss.

\subsection{Attention-Driven Post-hoc Interpretability with Adaptive Patch Clustering}

To interpret the classification output \(\hat{\mathbf y}\), we extract spatial attention maps from the final fusion layer. Let
\(
\mathbf A_{\text F\to\text L}^{(L)},\mathbf A_{\text L\to\text F}^{(L)}\in\mathbb R^{B\times h\times N\times N}
\)
denote cross-view attention weights.
Patch importance for the frontal view is computed by aggregating attention received from lateral tokens:
\begin{equation}
\mathcal H^{\text F}=
\text{softmax}\Big(\sum_{h=1}^{h}\mathbf A_{\text L\to\text F}^{(L)}[b,h,:,:]\Big)
\end{equation}
which is reshaped to the image resolution to form the heatmap \(\mathcal H^{\text F}\in[0,1]^{H\times W}\). The lateral heatmap \(\mathcal H^{\text L}\) is computed analogously.

To quantify view contributions, we compute 2-player Shapley values~\cite{lundberg2017unified}. For pathology class \(c\),
\begin{equation}
\phi_{\text F}^{(c)}=\frac12\Big(v(\{\text F\})-v(\emptyset)+v(\{\text F,\text L\})-v(\{\text L\})\Big)
\end{equation}
and similarly for \(\phi_{\text L}^{(c)}\). These are converted to percentage contributions
\(
\pi_{\text F}^{(c)},\pi_{\text L}^{(c)}.
\)

High-attention regions are then grouped into clusters. The heatmap is divided into a \(p\times p\) grid, producing patch scores \(s_{i,j}^v\). The top \(k\%\) patches are selected and clustered using DBSCAN~\cite{10.5555/3001460.3001507}:
\(
\mathcal C^v=\text{DBSCAN}(\mathcal Q^v,\varepsilon,m),
\)
where \(\varepsilon\) is the neighborhood radius and \(m\) is the minimum cluster size. Finally, the number of patches extracted from each view is dynamically allocated according to the Shapley percentages:
\begin{equation}
n_{\text F}^{(c)}=
\left\lfloor N_{\text total}\frac{\pi_{\text F}^{(c)}}{100}\right\rfloor,\quad
n_{\text L}^{(c)}=N_{\text total}-n_{\text F}^{(c)}.
\end{equation}
This produces view-adaptive patch sets that highlight diagnostically relevant regions.

\subsection{Hierarchical Cluster-Guided Visual Tokenization for Summarization Task}

Although attention heatmaps and clustered patches provide strong cues for pathology classification, directly concatenating these patches for report summarization introduces two limitations. 
First, independent patch extraction discards important structural information such as spatial position, cluster membership, and source view. Second, patches selected solely based on classification saliency may not be optimal for the downstream summarization task. 
To address these issues, we construct a hierarchical visual token representation that preserves structural context and dynamically supplements classification-driven patches with additional tokens relevant for text generation.

The attention heatmaps and clusters \(\mathcal C^{\text F},\mathcal C^{\text L}\) obtained from the interpretability stage are therefore used to construct structured visual tokens for the downstream T5-based report summarizer.
Let \(I^{\text F},I^{\text L}\) denote the \(224\times224\) frontal and lateral images. A ViT encoder~\cite{dosovitskiy2020image} produces
\(
\mathbf Z_{\text img}^{(v)}=\mathcal V(I^{(v)})\in\mathbb R^{B\times197\times D_{\text vit}},
\)
where 196 tokens correspond to the \(14\times14\) spatial grid. These tokens are projected into the T5 embedding space
\(
\mathbf E_{\text img}^{(v)}\in\mathbb R^{B\times197\times D_{\text t5}}, \quad D_{\text t5}=512.
\)
For each view \(v\), a hierarchical visual token sequence is constructed.

\textbf{Global token.}
A global representation is obtained from the CLS token and augmented with learnable view and token-type embeddings:
\begin{equation}
\mathbf t_{\text global}^{(v)}=
\text{CLS}^{(v)}+\mathbf e_{\text view}^{(v)}+\mathbf e_{\text type}^{(0)}.  
\end{equation}

\textbf{Cluster tokens.}
For each cluster \(c\in\mathcal C^{(v)}\), patch embeddings are aggregated using attention weights. Cluster geometry features (centroid, extent, size, and mean attention) are encoded through a small MLP to produce \(\mathbf e_{\text geom},c\). The cluster token is defined as
\begin{equation}
\mathbf t_c^{(v)}=
\Big(\sum_i w_i\mathbf E_{\text img,i}^{(v)}\Big)+
\mathbf e_{\text geom},c+
\mathbf e_{\text view}^{(v)}+
\mathbf e_{\text type}^{(1)}.    
\end{equation}
This representation preserves cluster-level structure while providing intermediate semantic context between local patches and global image tokens.

\textbf{Dynamic extra patches.}
To reduce over-reliance on classification-driven patches, remaining token capacity is filled using two lightweight scoring modules. 
An image-aware selector scores candidate patches using the global image token, while a text-aware selector additionally conditions on mean-pooled textual findings embeddings. 
These modules identify additional patches relevant to the summarization objective.

Each view produces a fixed set of n tokens, including global, cluster, and selected patch tokens. Frontal and lateral tokens are concatenated to form the visual sequence
\(
\mathbf E_{\text vis}\in\mathbb R^{B\times128\times D_{\text t5}}.
\)
Given textual findings embeddings \(\mathbf E_{\text text}\), the multimodal representation becomes
\(
\mathbf E_{\text fused}=[\mathbf E_{\text text};\mathbf E_{\text vis}]
\)
with corresponding attention mask \(\mathbf M_{\text fused}\). The T5 decoder then generates the final impression using teacher forcing and cross-entropy loss.

% \section{Experiment and Evaluation}
\section{Experiment and Result}

\subsection{Experimental Setup}
\subsubsection{Dataset Description}
The MIMIC-CXR dataset \cite{johnson2019mimiccxr_scidata} is a large-scale radiology report summarization benchmark comprising 377,110 chest X-ray images and 227,827 corresponding reports. We adopt an 8:1:1 split for training, validation, and testing.

\subsubsection{Evaluation Metrics.}
We evaluate generated summaries along two dimensions: similarity and factual correctness. For similarity, we use ROUGE-L F1 \cite{lin2004rouge}, along with BERTScore \cite{zhang2020bertscore}, BLEU-4 \cite{papineni2002bleu}, and METEOR \cite{banerjee2005meteor}. For factual correctness, we employ CheXbert \cite{smit2020chexbert} to compute label-based F1 scores and RadGraph \cite{jain2021radgraph} to assess clinical entity and relation accuracy.

Refer to \autoref{sec:impl} for implementation details.

% \section{Results and Discussion}

\begin{table}[t]
\centering
\resizebox{0.5\textwidth}{!}{
\begin{tabular}{@{}lcccccc@{}}
\hline
\textbf{Model} & \textbf{B-4} & \textbf{R-L} & \textbf{BScore} & \textbf{CBert} & \textbf{RadG} \\
\hline
\hline
\multicolumn{7}{c}{\textbf{Text-Only Baselines}} \\
\hline

BioBART  
& 19.71 & 43.68 & 90.66  & 58.84 & 36.68 & 45.33 \\

T5
& 21.83 & 44.61 & 91.45  & 38.49 & 38.26 & 46.79 \\

\hline

\multicolumn{7}{c}{\textbf{Full-Image Multimodal Models (Single vs Multi-View)}} \\
\hline

ViT + T5 (Frontal View)  
& 17.71 & 52.30 & 92.05 & 59.90 & 44.70 & 53.45 \\

ViT + T5 (Lateral View)  
& 20.30 & 56.42 & 92.74 & 59.06 & 49.29 & 56.89 \\

ViT + T5 (Frontal + Lateral Views) 
& 24.11 & 61.09 & 93.54 & 63.62 & 54.69 & 61.30 \\

\hline

\multicolumn{7}{c}{\textbf{Region-Guided Multimodal Models (ROI-based Visual Context)}} \\
\hline

ROI (Segmented Regions)
& 25.13 & 58.04 & 92.97 & 55.61 & 51.08 & 56.81 \\

ROI Patches (Frontal View)
& 24.95 & 57.45 & 91.64 & 60.57 & 53.92 & 57.68 \\

ROI Patches (Lateral View)
& 25.79 & 58.36 & 91.63 & 62.57 & 54.92 & 57.08 \\

Random Patches (Frontal + Lateral Views)
& 25.35 & 58.55 & 92.67 & 62.91 & 54.97 & 58.22 \\

\textbf{ViTAS (Selective Informative Patches, Ours)}
& \textbf{29.25} & \textbf{69.83} & \textbf{95.61} & \textbf{70.57} & \textbf{63.92} & \textbf{61.68} \\

\hline
\end{tabular}
}

\caption{
Ablation study on MIMIC-CXR summarization evaluating 
(1) text-only models, 
(2) full-image multimodal inputs (single vs multi-view), 
(3) ROI-guided visual context, and selective patch-based visual evidence. 
Results show that selecting salient visual evidence is more effective than simply increasing the amount of visual input. The evaluation metrics include BLEU-4 (B-4), ROUGE-L (R-L), BERTScore (BScore), CheXbert score (CBert), and RadGraph score (RadG).
}

\label{tab:ablation}

\end{table}

\subsection{Rethinking Visual Input in Multimodal Radiology Summarization}

We challenge two long-standing assumptions in multimodal medical report summarization. Prior radiology-report studies have repeatedly shown that multimodal models add little value when transforming richly detailed FINDINGS (already written by radiologists with all key image information) into the concise IMPRESSION section, attributing the gap to failed cross-modal interactions between text and images. At the same time, a large body of work in medical document understanding has operated under the heuristic that ``the more visual input, the better,'' assuming that simply concatenating more text and images will monotonically improve performance. Our ablation study on MIMIC-CXR provides the first rigorous evidence that \emph{both assumptions, while influential, do not fully account for the role of visual information in report summarization}. Table~\ref{tab:ablation} directly tests these claims. 

We first establish strong text-only baselines with BioBART~\cite{yao2023biobart} and T5~\cite{raffel2020exploring}. Introducing full-image visual context with ViT+T5 consistently improves performance across all evaluation metrics: BLEU-4, ROUGE-L, BERTScore, CheXbert, and RadGraph, with gains of 7.69\% (frontal view alone), 11.81\% (lateral view), and 16.48\% when both views are combined-clearly demonstrating the value of multi-view radiology images over pure text and refuting the claim that multimodal models provide little benefit in the FINDINGS $\rightarrow$ IMPRESSION step.

However, simply increasing the amount of visual information does not guarantee further improvements, challenging the common ``more is better'' assumption. When using region-guided inputs, segmented ROIs provide a modest gain of BLEU-4 of 3.3\% over text-only baselines, while ROI patches-frontal (+2.42\%) and lateral (+2.12\%)-and even random patches from both views (+2.52\%) show little additional benefit compared to the full multi-view setting, including all metrics. The plateau highlights a key challenge in multimodal learning: uncurated or noisy visual regions can dilute cross-modal alignment rather than strengthen it, emphasizing that quality and relevance of visual input matter more than sheer volume.

Our proposed ViTAS model, which selectively extracts only the most informative patches via attention-guided selection, breaks this plateau decisively. It achieves the highest overall gain of 25.22\% over the text-only baseline and outperforms the strongest full-image multi-view baseline, delivering substantially larger gains especially on the primary metrics. Critically, the improvements are most pronounced precisely in the FINDINGS $\rightarrow$ IMPRESSION transformation-the very stage where all previous multimodal systems had failed. Unlike prior multimodal approaches that fed full uncurated images (or naive ROI patches) directly into LVLMs and domain-specific fusion layers-thereby suffering from visual noise and failing to learn meaningful cross-modal interactions during summarization. Our framework first isolates lung ROIs with ensemble MedSAM2, performs bidirectional spatial cross-attention across complementary views, and then dynamically selects only the highest-Shapley-attention patches. This eliminates irrelevant regions, enforces clinically grounded text-image alignment, and is what finally enables multimodal models to outperform strong text-only baselines and produce accurate IMPRESSIONS where every previous system plateaued.

\subsection{Comparison with State-of-the-Art Models}

Table~\ref{tab:comparison_results} presents the performance of text-only and multimodal models on the MIMIC-CXR test set. Text-only baselines achieve moderate performance, with BART obtaining the highest ROUGE-L (47.00\%) and Llama-3.1-8B achieving highest CheXbert score (69.03\%), highlighting the limitations of using text alone. Multimodal models consistently outperform text-only baselines. For example, Flan-T5 + ViT reaches 25.87\% BLEU-4 and 77.93\% CheXbert score, confirming that visual information enhances report summarization. Our ViTAS model achieves the best results with 29.25\% BLEU-4 and 69.83\% ROUGE-L, demonstrating that selectively leveraging salient visual evidence across multi-view images effectively guides medical report summarization.

\begin{table}[!t]
\centering
\resizebox{\columnwidth}{!}{
\begin{tabular}{lccccc}
\hline
\textbf{Model} & \textbf{B-4} & \textbf{R-L} & \textbf{BScore} & \textbf{CBert} & \textbf{RadG} \\
\hline
\multicolumn{6}{c}{\textbf{Text-Only Baselines}} \\
\hline
PG~\cite{sim-etal-2025-more}     & --&36.84 &-- & 54.44 & 26.72 \\
BART~\cite{sim-etal-2025-more}   & -- & 47.00 & -- & 65.39 & 40.90 \\
GSum~\cite{sim-etal-2025-more}   & -- & 42.98 & -- & 59.60 & 35.18 \\
WGSum~\cite{sim-etal-2025-more}  & -- & 41.55 & -- & 58.92 & 31.17 \\
Llama-3.1-8B~\cite{zhao2025simplified}&11.67&32.46&54.56&69.03&29.16\\
Mistral-7B~\cite{zhao2025simplified}&8.67&29.81&51.58&66.33&24.91\\
Gemma-2-9b-it~\cite{zhao2025simplified}&10.16&30.71&52.95&68.42&27.44\\
\hline
\multicolumn{6}{c}{\textbf{Multimodal Baselines}} \\
\hline
Flan-T5 + ViT~\cite{wang-etal-2023-utsa-nlp} & 25.87 & 47.86 & 64.74 & 77.93 & 51.84 \\
OFA + ResNet~\cite{kim-etal-2023-ku-dmis-msra} & 25.58 & 47.75 & 64.80 & 76.29 & 50.96 \\
BLOOMZ-7b~\cite{karn-etal-2023-shs-nlp} & 25.32 & 47.48 & 63.61 & 74.34 & 49.00 \\
BioBART~\cite{wu-etal-2023-knowlab} & 22.97 & 46.15 & 63.43 & 75.14 & 48.04 \\
BERT + CvT-21~\cite{nicolson-etal-2023-e-health-csiro} & 17.97 & 44.14 & 61.47 & 71.67 & 44.95 \\
VG-BART (MHA)~\cite{sim-etal-2025-more} & -- & 46.89 & -- & 65.33 & 41.10 \\
VG-BART (Dot)~\cite{sim-etal-2025-more} & -- & 46.68 & -- & 65.59 & 40.92 \\
LLaVA-1.5~\cite{sim-etal-2025-more} & -- & 44.32 & -- & 62.14 & 38.81 \\
Qwen-VL~\cite{sim-etal-2025-more} & -- & 44.30 & -- & 63.08 & 38.83 \\
DeepSeek-VL~\cite{sim-etal-2025-more} & -- & 43.12 & -- & 62.95 & 37.79 \\
LLaVA-Med~\cite{sim-etal-2025-more} & -- & 44.53 & -- & 62.98 & 38.81 \\
\textbf{ViTAS (Ours)}
& \textbf{29.25} & \textbf{69.83} & \textbf{95.61} & \textbf{} & \textbf{}  \\
\hline
\end{tabular}
}
\caption{Performance comparison of SOTA models on the MIMIC-CXR test set. Text-only results are shown at the top, followed by multimodal models. }
\label{tab:comparison_results}
\end{table}

\subsection{Qualitative Analysis}

Table~\ref{tab:auroc_views} demonstrates that the Swin Transformer V2 backbone, which generates the attention heatmaps, achieves markedly higher AUROC on lateral views than on frontal views, with the largest gains on edema, pleural effusion, consolidation, and cardiomegaly. This confirms that lateral projections supply complementary diagnostic cues, directly motivating the bidirectional spatial cross-attention and Shapley-value view weighting in our multi-view fusion stage.

\begin{figure*}[htbp]
\centering
\includegraphics[width=0.7\textwidth]{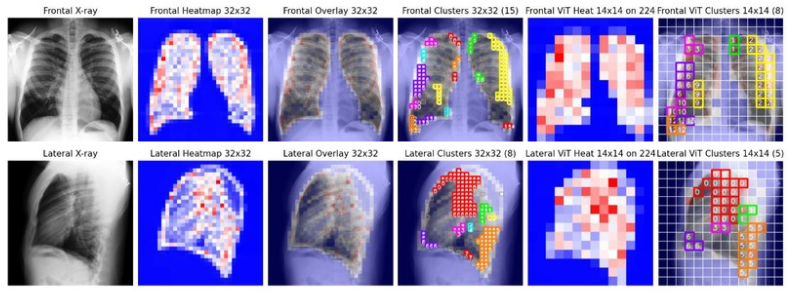}
\caption{Attention-driven patch selection pipeline (frontal top row, lateral bottom row): original X-ray, Swin V2 32$\times$32 heatmap (downsampled from 64$\times$64), overlay on MedSAM2-segmented lungs, DBSCAN clusters (15 frontal, 8 lateral), projected ViT 14$\times$14 heatmap, and final selected clusters (8 frontal, 5 lateral). Only these pathology-dominant patches feed the multimodal T5 decoder.}
\label{fig:attention_pipeline}
\end{figure*}

Figure~\ref{fig:attention_pipeline} shows the full attention-driven pipeline. MedSAM2 ensemble segmentation isolates the pulmonary fields. The Swin Transformer V2 produces 32$\times$32 heatmaps (downsampled from internal 64$\times$64 features), which are overlaid on the lung ROI. DBSCAN then forms spatially coherent clusters. Finally, heatmap scores are projected onto the 14$\times$14 ViT grid via overlap averaging, yielding the selected high-importance clusters. These steps eliminate background noise, convert scattered attention into anatomically meaningful groups, and ensure only the highest-Shapley-contribution patches reach the multimodal transformer.

\begin{figure}[!t]
    \centering
    \includegraphics[width=0.45\textwidth]{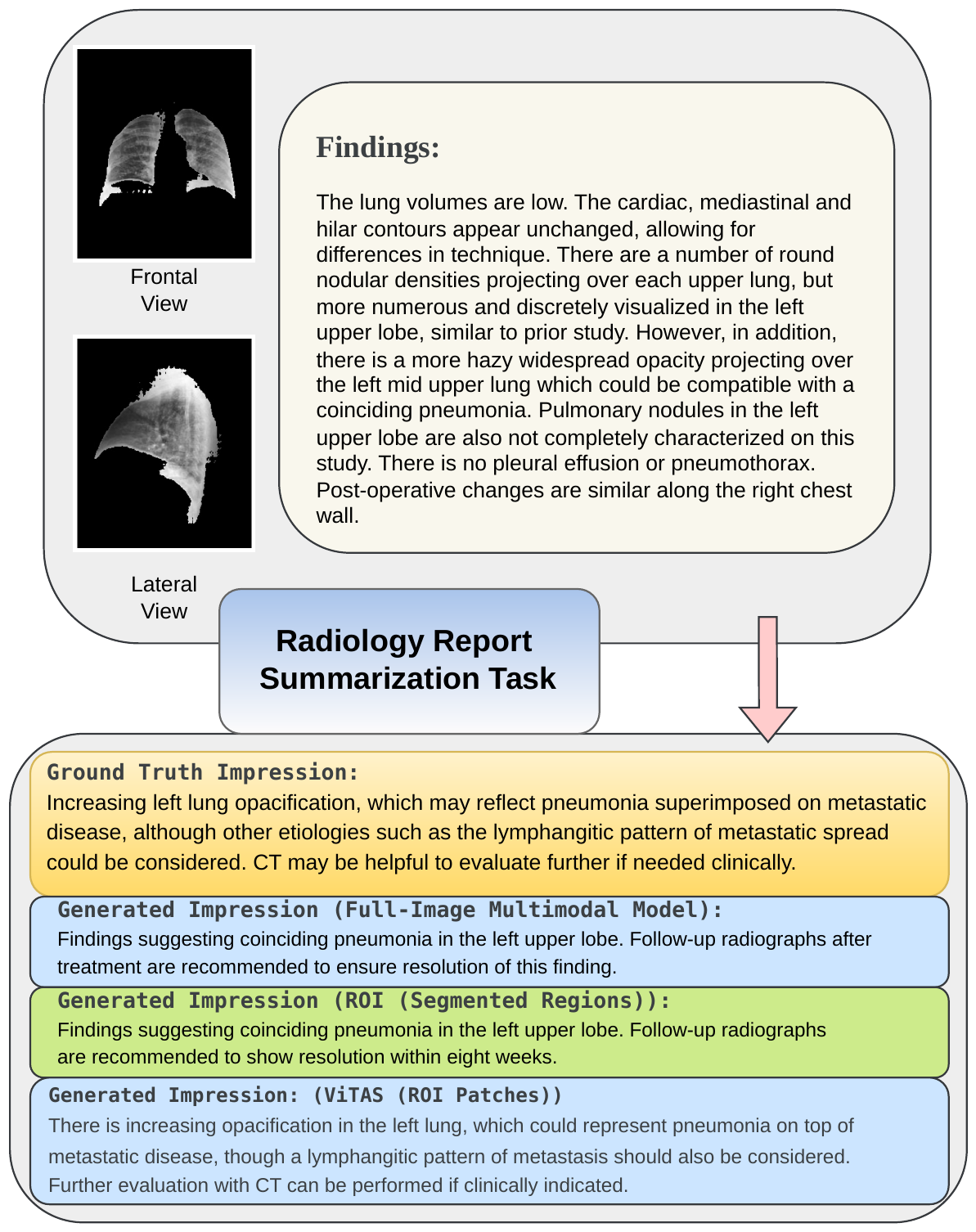} % replace with your image file
   \caption{Qualitative analysis of left lung opacification. Ground truth shows increasing opacification. Full-image and ROI models detect pneumonia with suggested follow-up, while the ViTAS ROI-patch model closely matches ground truth, capturing pneumonia and possible lymphangitic spread with suggestions.}
    \label{fig:lung_analysis}
\end{figure}

As shown in Figure~\ref{fig:lung_analysis}, the raw radiologist findings describe low lung volumes, stable cardiac contours, multiple round nodular densities (more numerous in the left upper lobe), and an additional hazy opacity in the left mid-upper lung suggestive of coinciding pneumonia, while ruling out effusion or pneumothorax. The full-image multimodal baseline correctly flags pneumonia but misses the metastatic nodular context entirely. The ROI-only model improves localization yet remains vague on the chronic-metastatic link. In contrast, our ViTAS approach (ROI patches selected via attention clustering and Shapley weighting) synthesizes the hazy opacity with the documented nodules into a precise impression. This closely matches the ground-truth expert interpretation while staying concise and actionable. ViTAS therefore outperforms the ablated variants because it feeds the T5 decoder exclusively with tokens from pathology-dominant, view-adaptive clusters, removing noise that dilutes both full-image and full-ROI baselines. The resulting impression is more faithful to the radiographic findings and clinically more useful-exactly the objective of our multi-stage framework.
\begin{table}[!t]
\centering
\resizebox{0.8\columnwidth}{!}{
\begin{tabular}{lcccc}
\hline
\textbf{Class} & \textbf{AP} & \textbf{PA} & \textbf{L} & \textbf{LL} \\
\hline
Atelectasis        & 0.7191 & 0.7227 & 0.8340 & 0.8128 \\
Cardiomegaly       & 0.8252 & 0.8162 & 0.9035 & 0.8507 \\
Consolidation      & 0.7509 & 0.7452 & 0.8485 & 0.8506 \\
Edema              & 0.8689 & 0.8713 & 0.9348 & 0.9128 \\
No Finding         & 0.7854 & 0.7857 & 0.8308 & 0.8241 \\
Pleural Effusion   & 0.8666 & 0.8650 & 0.9315 & 0.9568 \\
Pneumothorax       & 0.8447 & 0.8236 & 0.8956 & 0.7953 \\
Support Devices    & 0.7930 & 0.7888 & 0.7860 & 0.6941 \\

\textbf{Avg. AUROC} & \textbf{0.8067} & \textbf{0.8023} & \textbf{0.8706} & \textbf{0.8372} \\
\hline
\end{tabular}
}
\caption{Per-class AUROC across different chest X-ray projection views. AP, PA, L, and LL indicate Anteroposterior, Posteroanterior, Lateral, and Left Lateral views.}
\label{tab:auroc_views}
\end{table}

\begin{table}[!t]
\centering

\resizebox{\columnwidth}{!}{%
\begin{tabular}{l c c c c c}
\hline
Model & Readability & Factual Correctness & Informativeness & Redundancy & Completeness \\ \hline
Full-Image Multimodal Model & 4.11 & 4.22 & 3.78 & 4.00 & 3.67 \\
ROI (Segmented Regions) & 4.15 & 4.18 & 3.85 & 4.05 & 3.72 \\
ViTAS (ROI Patches) & 3.92 & 5.00 & 4.95 & 3.88 & 5.00 \\ \hline
\end{tabular}%
}
\caption{Human evaluation results based on summarization quality. Higher scores indicate better performance. Scores are reported as floats with two decimal points.}
\label{tab:qualitative_scores}
\end{table}

\paragraph{Human Evaluation}
We conducted expert evaluations to correlate model-generated summaries with human judgment. A total of 100 randomly selected samples from the dataset were reviewed. Seven volunteers participated, including one MBBS doctor, two final-year medical students, two radiology researchers, and two clinical radiologists. Each participant rated the summaries on a scale from 1 (very poor) to 5 (very good) across five dimensions: readability, factual correctness, informativeness, redundancy, and completeness. Table~\ref{tab:qualitative_scores} shows that all three models performed closely, with the ViTAS model using ROI patches achieving slightly higher scores in most categories. Upon inspection, we found that using ROI patches allowed the model to better capture key clinical details, and maintain factual correctness, resulting in improved alignment with expert judgment.

\section{Conclusion}
We presented ViTAS, a selective-attention multimodal framework that challenges the conventional ``more is better'' paradigm in radiology report summarization. By systematically removing visual noise through ensemble segmentation, bidirectional cross-view fusion, and pathology-adaptive patch selection, ViTAS achieves SOTA performance on MIMIC-CXR while outperforming both full-image multimodal and strong text-only baselines. Qualitative analysis and human evaluation further confirm superior factual alignment, reduced hallucination, and better clinical utility. These results suggest that future multimodal medical summarization systems should prioritize relevance over volume in visual input. This work would inspire more interpretability-driven and noise-aware approaches in clinical NLP and medical vision-language modeling.

\section*{Limitations}
Our approach relies on accurate lung segmentation and attention heatmaps from a frozen Swin Transformer V2 classifier; errors in either stage may propagate to patch selection.
The current implementation is evaluated only on MIMIC-CXR (single-institution, chest X-rays); generalization to multi-modal, multi-organ, or out-of-domain datasets remains untested.
% Human evaluation, while conducted with domain experts, used a relatively small sample (100 reports) and five raters. 
% Training ViTAS requires significant compute (NVIDIA A100 GPUs), and inference speed is slower than text-only models due to the multi-stage pipeline. 
% Finally, we do not explicitly constrain the model with external knowledge bases or rule-based fact-checking.

% \section*{Acknowledgement}
% We used generative AI only to enhance the quality of English, with all outputs carefully reviewed and verified by the authors.

\bibliography{references}

\appendix

\section{Implementations and Model Details.}\label{sec:impl}
We implement our model in PyTorch on an NVIDIA A100 GPU. The lung ROI extraction uses MedSAM2 with five bounding box prompts (one reference and four shifts of 32-48 pixels), followed by morphological closing with a $5\times5$ kernel. For the dual-view classifier, images are resized to $256\times256$ and processed using two Swin Transformer V2-Small backbones, with bidirectional cross-attention inserted after stage 3 (8 heads, embedding dimension 768). The model is trained for 100 epochs using the Ranger optimizer (learning rate $3\times10^{-4}$, weight decay 0.05), cosine annealing with 100 warmup epochs, and batch size 32. For interpretability, Grad-CAM-style backpropagation through the final cross-attention layer produces $32\times32$ attention maps, upsampled to full resolution, while Shapley estimation requires 4 forward passes per sample. DBSCAN clustering is applied with $\varepsilon=1.8$ and minPts = 4 on the top 25\% attention regions. In the summarization stage, ViT-B/16 encoders ($224\times224$) generate 64 tokens per view (128 total), projected to 512 dimensions and fused with text embeddings for a T5-small decoder. The model is trained for 100 epochs using Adafactor (learning rate $1\times10^{-3}$), batch size 32, label smoothing 0.1, and maximum target length 256.

\end{document}